\documentclass[conference]{IEEEtran}
\IEEEoverridecommandlockouts
\usepackage{cite}
\usepackage{amsmath,amssymb,amsfonts}
\usepackage{algorithmic}
\usepackage{graphicx}
\usepackage{textcomp}
\usepackage{xcolor}
\usepackage{url}
\usepackage{tabularx}
\def\BibTeX{{\rm B\kern-.05em{\sc i\kern-.025em b}\kern-.08em
    T\kern-.1667em\lower.7ex\hbox{E}\kern-.125emX}}
\begin{document}

\title{PUPILLOMETRY AND BRAIN DYNAMICS FOR COGNITIVE LOAD IN WORKING MEMORY\\

}

\author{\IEEEauthorblockN{1\textsuperscript{st} Nusaibah Farrukh}
\IEEEauthorblockA{\textit{Digital University Kerala} \\
Thiruvananthapuram, Kerala  \\
nusaibah.farrukh@gmail.com \\ORCID: 0009-0004-1237-2693}
\and
\IEEEauthorblockN{2\textsuperscript{nd} Malavika Pradeep}
\IEEEauthorblockA{\textit{Digital University Kerala} \\
Thiruvananthapuram, Kerala \\
malavikapradeep2001@gmail.com \\ ORCID: 0009-0005-9215-3542}
\and
\IEEEauthorblockN{3\textsuperscript{rd} Akshay Sasi}
\IEEEauthorblockA{\textit{Digital University Kerala} \\
Thiruvananthapuram, Kerala \\
akshaysasi12.knr@gmail.com \\ ORCID: 0009-0009-3708-554X}
\and
\IEEEauthorblockN{4\textsuperscript{th} Rahul Venugopal}
\IEEEauthorblockA{\textit{Centre for Consciousness Studies, NIMHANS} \\
Bangalore, India \\
rhlvenugopal@gmail.com \\ ORCID: 0000-0001-5348-8845}
\and
\IEEEauthorblockN{5\textsuperscript{th} Elizabeth Sherly}
\IEEEauthorblockA{\textit{Digital University Kerala} \\
Thiruvananthapuram, Kerala \\
sherly@duk.ac.in \\ ORCID: 0000-0001-6508-950X }
}

\maketitle

\begin{abstract}
Cognitive load, the mental effort required during working memory, is central to neuroscience, psychology, and human computer interaction. Accurate assessment is vital for adaptive learning, clinical monitoring, and brain–computer interfaces. Physiological signals such as pupillometry and electroencephalography (EEG) are established biomarkers of cognitive load, but their comparative utility and practical integration as lightweight, wearable monitoring remain underexplored. 

EEG provides high temporal resolution of neural activity. Its non invasive but is difficult in technology and suffers from limitations of wearability and cost as it is resource-intensive, whereas pupillometry is non invasive, portable, and scalable. Existing studies usually depend on deep learning models with little interpretability and substantial computational expense. This research extends recent progress in time-series analysis by synthesizing feature based and model driven methodology. This study builds on recent advances in time-series analysis by integrating feature based and model driven approaches. 

Using the OpenNeuro “Digit Span Task” dataset, this study investigates classification of cognitive load levels from EEG and pupillometry. Feature based approaches with Catch-22 features and classical machine learning models outperform deep learning in both binary (just listen vs. memory) and multiclass (varying memory loads) tasks.The findings demonstrate that pupillometry by itself can compete with EEG, providing a portable, useful proxy for real-world implementation. 

These findings challenge the assumption that EEG is necessary for load detection by showing that pupil dynamics, in conjunction with interpretable models and feature analysis based on SHAP, offer physiologically significant insights. This creates a path for wearable, affordable cognitive monitoring with uses in neuropsychiatry, education, and healthcare by developing interpretable multimodal models.

\end{abstract}


\section{Introduction}
Cognitive load in working memory is a primary concern in education, ergonomics, and clinical science, and a constrained system for temporary processing and storage. Cognitive load refers to the mental effort involved in maintaining and manipulating information in working memory. Two systems are relied upon by working memory performance: central executive for attentional control and a short-term buffer for storage. Overload happens when demands posed by tasks surpass this capacity, such
as remembering 13 digits compared to five. Digit span tests, long employed in neurosciences and clinical trials, are still a common method to measure memory and attention under different cognitive load \cite{b6}.

\begin{figure}[htbp]
  \centering
  \includegraphics[width=0.4\textwidth]{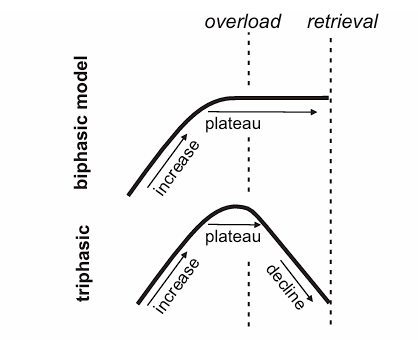}  
  \caption{Biphasic and Triphasic Models of Pupil Response to Increasing Cognitive Load \cite{b1}}
  \label{fig:pupil_response}
\end{figure}

Electroencephalography (EEG) and pupillometry are two modalities that have been identified as particularly informative.  A non-invasive indicator of cognitive effort that is controlled by sympathetic and parasympathetic pathways is pupillometry, which measures pupil size and reactivity.  Numerous studies that use working memory tasks or digit span have found that, in the context of cognitive load, pupil size increases linearly as the number of memory items increases up to the individual's cognitive capacity (e.g., Karatekin, 2004 \cite{b2}; Klingner et al., 2011 \cite{b3}; Piquado et al., 2010 \cite{b4}).  However, as Figure~\ref{fig:pupil_response} illustrates, the pupil’s response to increasing cognitive demands is not strictly linear. 

In addition, EEG offers high temporal resolution of neural dynamics, with working memory load being associated with frontal theta and parietal alpha rhythms [5]. EEG is frequently used in cognitive neuroscience research, brain-computer interfaces, and clinical diagnosis (such as epilepsy and sleep disorders), offering information on perception, attention, memory, and cognitive load. EEG and pupillometry represent complementary approaches: one probing neural dynamics directly, the other capturing an accessible physiological proxy. Thus the paper examines the interaction between brain activity and pupil dynamics under varying cognitive demands.

Traditional approaches to measuring cognitive load rely on behavioral tasks, self-report measures, and neuroimaging techniques. EEG remains resource intensive and impractical outside labs. Behavioral and self-report methods, meanwhile, fail to capture real-time fluctuations. The unresolved challenge is how to combine EEG and pupillometry into practical, interpretable systems for cognitive load monitoring. Unlike EEG or fMRI, pupillometry requires minimal equipment and can be implemented using wearable, spectacle-mounted eye trackers, making it a promising candidate for real-world cognitive monitoring.

This study addresses the problem by introducing a unified pipeline for preprocessing, feature extraction (via Catch-22), and classification using both EEG and pupillometry. Leveraging the OpenNeuro Digit Span dataset (64 participants, tasks with 5, 9, or 13 digits plus passive listening), we perform binary classification (just listen vs. memory) and multiclass classification (graded load) to investigate how cognitive load can be decoded from physiological signals. Models range from Random Forest, XGBoost, and SVM to CNNs, LSTMs, and Transformers, with SHAP-based feature importance applied for interpretability.

Findings show that pupillometry is a strong proxy for EEG, especially with feature-based models. While feature-engineered methods outperformed deep neural networks, which were slower and less interpretable, XGBoost and Random Forest consistently achieved high accuracy, frequently matching EEG.  Pupil dilation patterns and EEG theta/alpha activity are examples of physiologically believable predictors that SHAP validated.  In conclusion, this work shows that cognitive load can be reliably classified using wearable, lightweight signals and advances cognitive load research towards practicality.  This research is important not only for its scientific contributions but also because it advances scalable and interpretable cognitive state assessment and opens the door to real-world applications in neuropsychiatry, education, elder care, and clinical monitoring.

\section{Related Works}

Human–computer interaction and cognitive neuroscience research has increasingly concentrated on evaluating cognitive load in real time using physiological signals like pupillometry and EEG. Reliable measures of mental effort are captured by both modalities, and efficient classification of cognitive states is made possible by machine learning and signal processing techniques \cite{b10}.

During digit recall tasks, Kosachenko et al.  \cite{b1} demonstrated that posterior alpha followed a different trajectory, whereas pupil size and frontal midline theta tracked cognitive load in parallel. Pupil dilation, which usually increases with load until working memory capacity is exceeded, is also shown to reflect cognitive control effort rather than task difficulty, according to studies \cite{b11}. EEG-based research also emphasises how parietal/occipital alpha suppression and frontal theta increases reflect increased cognitive load [12]. 

Multimodal approaches have been proposed, such as fusing EEG and pupil area signals for depression detection, achieving significant improvements in classification accuracy \cite{b17}. Pupillometry has also been analyzed as a psychophysiological tool, with the Psychosensory Pupil Response (PPR) reflecting arousal and mental effort via the locus coeruleus–norepinephrine system \cite{b13}, \cite{b14}, \cite{b15}. These studies emphasize the importance of standardized experimental designs and pre-processing methods for reliable pupil-based measures.

Advances in deep learning have expanded EEG applications, with transformer architectures outperforming traditional models in tasks like motor imagery and emotion recognition \cite{b16}. Multimodal studies combining fMRI, EEG, and pupillometry further demonstrate improved classification of mind-wandering states compared to unimodal approaches \cite{b18}. Feature based approaches such as catch22 also provide efficient, interpretable time-series representations applicable across diverse datasets \cite{b8}.

Overall, prior work has established the utility of EEG and pupillometry for cognitive load detection, but challenges remain regarding small datasets, modality alignment, and model interpretability.

\section{Methodology}
\begin{figure}[htbp]
  \centering
  \includegraphics[width=0.5\textwidth]{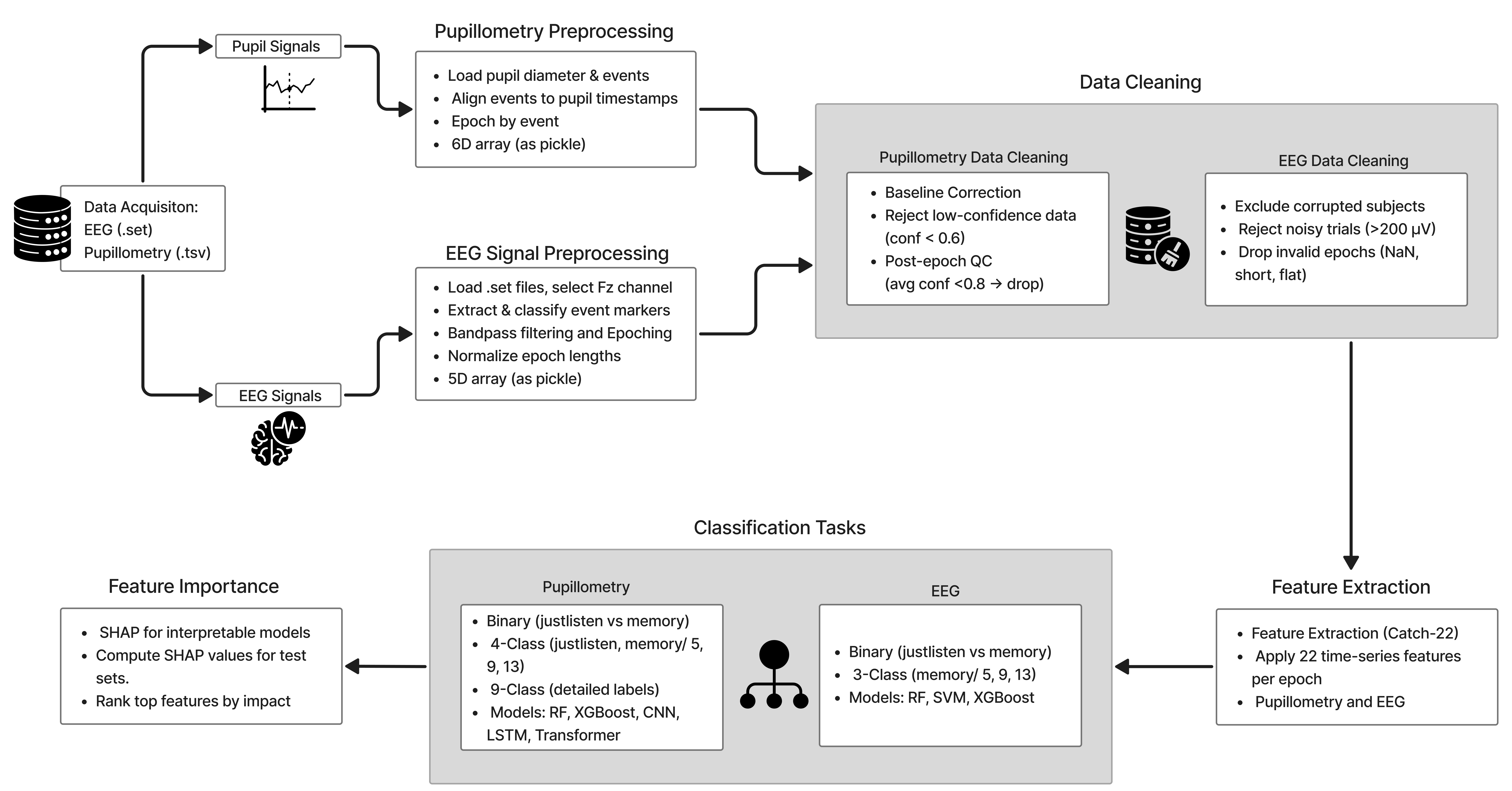}  
  \caption{Flowchart outlining Methodological Pipeline.}
  \label{fig:Flowchart}
\end{figure}

We employed multimodal physiological data from the OpenNeuro Digit Span Task dataset to investigate cognitive load through EEG and pupillometry. A structured pipeline was implemented as shown in Figure~\ref{fig:Flowchart}, covering acquisition, preprocessing, cleaning, feature extraction, classification, and interpretability for both EEG and pupillometry signals.

\subsection{Data acquisition \& Task paradigm}\label{AA}
\begin{figure}[htbp]
  \centering
  \includegraphics[width=0.4\textwidth]{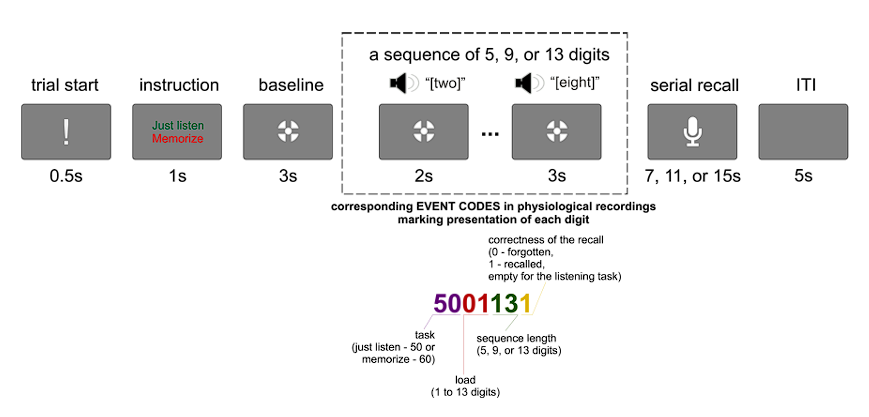}  
  \caption{Experimental design of the Digit Span working memory task}
  \label{fig:taskdescription}
\end{figure}

The dataset \cite{b7} comprises raw 64-channel EEG, cardiovascular measures (ECG and PPG), pupillometry, and behavioral responses from 86 participants.The dataset is available on: https://openneuro.org/datasets/ds003838 and is BIDS-formatted with per-subject folders (eeg/.set, eyetrack/.tsv), plus participants.tsv, events.tsv and JSON metadata. The analysis uses the subset of 64 participants reported in this work; only EEG and pupillometry modalities were taken forward. Participants performed a digit-span task after a 4-minute eyes-closed rest; each trial began with a cue, a 3 s baseline, then a sequence of 5, 9 or 13 digits followed by a retention/recall period as shown in Figure~\ref{fig:taskdescription}. Pupillometry used a Pupil Labs wearable tracker (one-point calibration per session) at 120 Hz and EEG was recorded with a 64-channel ActiCHamp system (extended 10–20, FCz reference, Fpz ground) at 1000 Hz. 

We performed Exploratory Data Analysis on participants.tsv for demographics and missing values (no critical inconsistencies) and checked eye-dominance, handedness and basic age/gender distributions to inform later stratification. Subjective workload (NASA-TLX) trends were summarized to provide context for physiological responses (mental demand high, small fatigue trends across blocks).

\subsection{Signal Pre-processing}
\subsubsection{Pupillometry Preprocessing}
Files were loaded per subject and per eye, events filtered to memory/control conditions, and event timestamps aligned to the nearest pupil timestamp. Custom event codes (e.g., \texttt{justlisten/five: 500105}; \texttt{memory/correct/five: 6001051}) were used and epoch windows were set per load: 5-digit: $-3 \rightarrow +10$ s; 9-digit: $-3 \rightarrow +18$ s; 13-digit: $-3 \rightarrow +26$ s. Epoched data were aggregated into a structured 6-D NumPy array (subjects $\times$ conditions $\times$ epochs $\times$ timepoints $\times$ features $\times$ eyes) and persisted as a pickle for downstream steps.

\subsubsection{EEG Preprocessing}
Only frontal channel Fz was retained (chosen for attentional/frontal activity). Epochs used $-3$ s pre-event to $+5$/$+15$/$+23$ s depending on condition; baseline correction was intentionally omitted to retain raw waveform features. Signals were normalized in length by trimming or zero-padding (\texttt{NaN}) to a consistent maximum length (26,000 samples) for analysis. Processed data were structured as a 5D array (subjects $\times$ conditions $\times$ epochs $\times$ time $\times$ channels).

\subsubsection{ Data Cleaning}
Pupillometry: Each pupil sample carried a confidence score provided by the eye-tracker, reflecting the reliability of gaze detection (e.g., whether the pupil was clearly visible, not occluded by blinks or motion). Low scores usually indicate signal loss, occlusion, or poor tracking quality.

Samples with confidence $< 0.6$ were removed to ensure high-quality data, and epochs with an average confidence $< 0.8$ were entirely discarded. This thresholding removes untrustworthy measurements, thereby reducing, as much as possible, artifacts from blinks or noise, improving the validity of downstream analyses. Dropped epochs were recorded by subject and condition. Quality checks were implemented to check proper alignment of event markers with pupil timestamps to avoid temporal drift.

EEG: Rejected trials were those above a peak-to-peak voltage of 200~$\mu$V (more commonly eye blinks, muscle artifacts, or electrode pops), flatlined with little variance (representing dropout of signal), or containing NaNs because of recording errors. Entire subjects with bad or missing files were removed. These exclusions guaranteed that only biologically meaningful and technically sound EEG data went into analysis.

\subsection{Feature extraction and Interpretability}
To obtain compact and interpretable representations, we extracted Catch-22 (Canonical Time-series Characteristics) features from both EEG and pupillometry signals. Catch-22 provides 22 minimally redundant descriptors capturing statistical, spectral, and dynamical properties of time series, offering dimensionality reduction while preserving discriminative information \cite{b8}. Features captured distributional, autocorrelation, spectral, and fluctuation dynamics.

For pupillometry features were computed per epoch and per eye. Features across eyes were either averaged or concatenated to form the final input set and for EEG, Catch-22 features were extracted from the Fz channel for each epoch. The feature set ensured interpretability and computational efficiency across classification tasks.

SHAP (SHapley Additive exPlanations) was used with tree-based models to quantify feature contributions, highlighting temporal complexity and entropy measures as key predictors. Deep learning models were excluded from SHAP analysis due to computational constraints.
For each trained model, SHAP values were calculated on the test set, and the top 10 features were identified by summing absolute SHAP values across all samples. This ranking highlighted recurrent patterns, such as temporal complexity, entropy, and fluctuation-related features, as the most influential in distinguishing cognitive load states.

\subsection{Classification}

\textit{Tasks:}Four classification tasks were defined:

1. Binary – just listen vs memory.

2. 4-class(pupillometry) – just listen, memory/5, memory/9, memory/13.

3. 3-class(EEG) -  memory/5, memory/9, memory/13.

4. 9-class – just listen, memory/correct, and memory/incorrect across three loads.

\textit{Models:} Random Forest (RF), XGBoost, SVM, 1D-CNN, LSTM (3-layer bidirectional with attention components), and a lightweight Transformer (1 encoder layer, 2 attention heads). Key hyperparameters used in experiments include RF (up to 1000 trees for ensembles), XGBoost ($\approx 100$ trees), and class-balancing-aware training for deep models (class weights, early stopping, ReduceLROnPlateau, Adam/AdamW optimizers).

Oversampling strategies applied on training data included SMOTE, SMOTE-ENN, and ADASYN (selected per task to mitigate class imbalance). Evaluation used stratified splits and metrics: accuracy, precision, recall, F1, confusion matrices (ROC curves for binary/EEG analyses). Tables \ref{tab:method_comparison} and \ref{tab:hyperparameters} summarize the classification approaches, computational complexity, and model hyperparameters.

\begin{table}[ht]
  \centering
  \caption{Summary of Classification Approaches and Computational Complexity}
  \label{tab:method_comparison}
  \renewcommand{\arraystretch}{1.2}
  \footnotesize 
  \begin{tabular}{|p{1.5cm}|p{1.2cm}|p{1.5cm}|p{1.9cm}|}
    \hline
    \textbf{Method Type} & \textbf{Models} & \textbf{Computation Complexity} & \textbf{Approx. Resource Usage} \\
    \hline
    Classical ML & Random Forest, XGBoost, SVM & Low & $\sim$2--4 GB RAM, CPU sufficient \\
    \hline
    Deep Learning & CNN, LSTM & Medium & $\sim$6--12 GB RAM, GPU recommended \\
    \hline
    Transformers & Lightweight Transformer & High & $\sim$12--16 GB RAM, GPU strongly recommended \\
    \hline
  \end{tabular}
\end{table}

\begin{table}[ht]
\centering
\caption{Model Hyperparameters}
\renewcommand{\arraystretch}{1.2}
\begin{tabularx}{\linewidth}{|l|X|}
\hline
\textbf{Model} & \textbf{Hyperparameters} \\
\hline
Random Forest (RF) & Number of trees: 100, 500, 1000; Max depth: None; Criterion: Gini; Class weighting applied \\
\hline
XGBoost & Trees: 100; Max depth: 5; Learning rate: 0.1; Subsample: 0.8; Colsample\_bytree: 0.8 \\
\hline
SVM & Kernel: RBF; $C=1.0$; Gamma = ``scale''; Regularization parameter C tuned; Class weights balanced \\
\hline
CNN & 1D Conv layers: 2; Filters: [32, 64]; Kernel size: 3; Dropout: 0.5; Optimizer: Adam (lr=0.001); ReLU activations; Max-pooling; Epochs: 50; Batch size: 32 \\
\hline
LSTM & Bidirectional LSTM layers: 3; Hidden units: 64; Attention layer; Dropout: 0.5; Optimizer: Adam (lr=0.001); Epochs: 50; Batch size: 32; Early stopping \& ReduceLROnPlateau applied \\
\hline
Transformer & Encoder layers: 1; Attention heads: 2; Hidden size: 128; Feedforward dim: 256; Dropout: 0.1; Optimizer: AdamW (lr=0.0005); Epochs: 50; Batch size: 32; ReduceLROnPlateau; Early stopping \\
\hline
\end{tabularx}
\label{tab:hyperparameters}
\end{table}

\section{Results}
\subsection{Pupillometry-Based Classification}
In the binary task, LSTM achieved the highest accuracy at 66.7\% but had the longest training time of 902.48~s. XGBoost provided the best efficiency--performance balance, attaining an accuracy of 61.3\%, a weighted F1(W) of 0.55, and the fastest training and prediction times of 1.45~s and 0.02~s, respectively.

CNN achieved a comparable weighted F1 of 0.56 but was roughly $280\times$ slower than XGBoost in terms of training. The Transformer model underperformed, showing 44.6\% accuracy with a reasonable training cost of 134.85~s. From the results shown in Table 3, it is evident that although the sequence models perform temporal feature extraction on pupil dynamics, the tree-based models are computationally more scalable for real-time deployment.

In the 4-class task, XGBoost again emerged as the best performer, achieving an accuracy of 52.3\% and the highest macro-F1(M) score of 0.53. Random Forest performed considerably worse, with an accuracy of 42.3\%, while CNN obtained 41.4\%. All models misclassified the \textit{Just Listen} class consistently, yielding F1~$<0.20$. 

Conversely, the \textit{Memory-5} condition showed the highest detection reliability across models, with precision/recall~$>0.50$. When trained directly on raw pupil sequences, XGBoost achieved an accuracy of 52.8\%, whereas CNN and LSTM dropped to near-chance performance, demonstrating that engineered features such as the Catch-22 set improve class separability by more than 25\% relative to raw-sequence models.

\begin{table}[htbp]
\caption{Model performance comparison of pupillometry data in binary and multiclass classification}
\centering
\renewcommand{\arraystretch}{1.35}
\setlength{\tabcolsep}{2pt} 
\normalsize
\begin{tabularx}{\linewidth}{|l|X|X|X|X|X|X|}
\hline
\textbf{Model} & \textbf{Binary Acc.} & \textbf{Binary F1 (W)} & \textbf{Binary F1 (M)} & \textbf{Multi Acc.} & \textbf{Multi F1 (W)} & \textbf{Multi F1 (M)} \\
\hline
XGBoost       & 0.613 & 0.55 & 0.45 & 0.524 & 0.50 & 0.53 \\
Random Forest & 0.326 & 0.23 & 0.29 & 0.423 & 0.38 & 0.43 \\
CNN           & 0.583 & 0.56 & 0.49 & 0.414 & 0.33 & 0.37 \\
LSTM          & 0.667 & 0.53 & 0.40 & 0.462 & 0.37 & 0.41 \\
Transformer   & 0.446 & 0.46 & 0.44 & 0.435 & 0.40 & 0.43 \\
\hline
\end{tabularx}
\label{tab:model_performance}
\end{table}

\subsection{EEG-Based Classification}

For binary EEG classification as showin in table 4, the best cross-validation(CV) accuracy ($72.3\% \pm 0.0076$) and the best test accuracy (57.3\%) were obtained using XGBoost. The Random Forest and SVM models yielded test accuracies of 56.6\% and 48.6\%, respectively. All models showed a systematic tendency to predict the \textit{Memory} class, where the false-negative rates for the \textit{Just Listen} class exceeded 35--40\% due to the underlying class imbalance in the dataset.

In the multiclass task as , all models achieved near-perfect performance. XGBoost reached a test accuracy of 99.8\% and an F1 score of 1.0, while Random Forest and SVM achieved 99.4\% and 98.0\%, respectively. The gap between cross-validation and test performance remained small ($<0.5\%$ for both XGBoost and Random Forest), indicating stable training. The results are presented in Table 5. 

However, uniformly high F1-scores (0.98--1.00) and extremely low error rates ($<0.5\%$) suggest potential overfitting or information leakage. As EEG data typically exhibits high inter-subject variability, these results should be interpreted with caution until replicated on larger or cross-subject datasets.

\begin{table}[!t]
\centering
\caption{Model performance comparison of EEG data in binary classification}
\label{tab:eeg_binary}
\small 
\setlength{\tabcolsep}{4pt}
\renewcommand{\arraystretch}{1.2}
\begin{tabular}{|l|c|c|c|}
\hline
\textbf{Model} &
\textbf{CV Acc.} &
\textbf{Test Acc.} &
\textbf{Macro F1} \\
\hline
Random Forest & $0.6156 \pm 0.0101$ & 0.5657 & 0.49 \\
SVM           & $0.5035 \pm 0.0072$ & 0.4860 & 0.47 \\
XGBoost       & $0.7230 \pm 0.0076$ & 0.5732 & 0.50 \\
\hline
\end{tabular}
\end{table}

\begin{table}[!t]
\centering
\caption{Model performance comparison of EEG data in multiclass classification}
\label{tab:eeg_multiclass}
\small
\setlength{\tabcolsep}{4pt}
\renewcommand{\arraystretch}{1.2}
\begin{tabular}{|l|c|c|c|}
\hline
\textbf{Model} &
\textbf{CV Acc.} &
\textbf{Test Acc.} &
\textbf{Macro F1} \\
\hline
Random Forest & $0.9975 \pm 0.0013$ & 0.9945 & 0.99 \\
SVM           & $0.9840 \pm 0.0059$ & 0.9799 & 0.98 \\
XGBoost       & $0.9977 \pm 0.0010$ & 0.9982 & 1.00 \\
\hline
\end{tabular}
\end{table}

\subsection{Feature Importance and Interpretability}
For both pupillometry and EEG data, feature extraction leveraged Catch-22 features, selected using SHAP analysis for their high relevance in marking cognitive load. The most critical features for pupillometry included: $DN\_HistogramMode\_5/10$ (capturing distributional shape and baseline vs. dilation shifts), $CO\_trev\_1\_num$ (indicating time-reversal asymmetry, relevant for dilation versus constriction slopes), $SB\_BinaryStats\_mean\_longstretch1$ (quantifying sustained load by the longest run of dilation/constriction), $MD\_hrv\_classic\_pnn40$ (a pupil variability, HRV-like measure), and $FC\_LocalSimple\_mean1\_tauresrat$ (describing local predictability, linked to irregularity under stress). These capture how pupil dilation dynamics reflect temporal regularity, baseline shifts, and asymmetry in constriction–dilation cycles under cognitive load. 

For EEG, the highly relevant features were: $DN\_HistogramMode\_5/10$ (distribution of amplitudes), $CO\_FirstMin\_ac$ (autocorrelation, relating to oscillatory patterns such as theta/alpha), $CO\_trev\_1\_num$ (asymmetry in time-series, relevant to ERP dynamics), $FC\_LocalSimple\_mean1\_tauresrat$ (predictability ratio, distinguishing structured from noisy oscillations), $SP\_Summaries\_welch\_rect\_area\_5\_1$ (a spectral power proxy for oscillatory energy) and $SB\_MotifThree\_quantile\_hh$ (identifying recurrent motifs potentially linked to ERP components). Collectively, these features provide physiologically interpretable markers of cognitive load by capturing distributional, autocorrelation, asymmetry, spectral, and variability properties of the time-series data.

\subsection{Discussion}
In general, XGBoost was the best balanced and most efficient model of all modalities, with high performance at low computational expense. Deep learn-ing models, particularly LSTM, were able to learn temporal dependencies in pupillometry but needed prohibitive amounts of resources and overfit. EEG performed significantly better than pupillometry with nearly perfect cognitive load level decoding however, practical application is limited by hardware needs, whereas pupil based monitoring is lightweight and scalable. 

A repeated limitation across modalities was the inability to detect the Just Listen state, which was probably owing to overlapping statistical patterns with memory tasks. This underscores the necessity of multimodal integration (e.g., EEG + pupillometry) or more complex feature extraction to more adequately represent low engagement states.
Lastly, while results were encouraging in controlled settings, deployment in the real world might encounter variability in lighting, sensor placement, and user state. Transfer learning, domain adaptation, and multimodal fusion are avenues of future work that must be investigated to enhance robustness in naturalistic environments.

Complementary research with the same OpenNeuro Digit Span dataset has additionally illustrated the utility of physiological signals for monitoring cognitive state. One study investigated heart–brain coupling, indicating that ECG derived Heart Rate Variability (HRV) and Catch-22 features can accurately predict cognitive load, with accuracies on par with EEG based models, and enabling the application of ECG as a convenient surrogate for neural indicators. Another parallel investigation created a crossmodal regression model associating ECG derived HRV aspects to EEG-based cognitive measures, and suggested synthetic HRV generation via the Poincar´e Sympathetic Vagal model in order to mimic brain–heart interactions. Collectively, these results affirm the general conclusion that multimodal physiological signals across EEG, pupillometry, and ECG provide complementary, scalable, and interpretable cognitive load biomarkers enabling wearable and real-time neuroergonomic applications.

\section{Conclusion and Future Work}

This work explored the multimodal classification of cognitive load from EEG and pupillometry using interpretable feature extraction and computationally lightweight models. A major contribution is the focus on field deployability, building models capable of running on low-power hardware, in contrast to current GPU-intensive deep learning methods. Experimental results demonstrate reliable pupillometry and EEG signal decoding of cognitive load. 

Feature-based methods, especially Catch-22, enhanced classification performance and interpretability, with traditional machine learning models performing better on raw pupillometry data than CNN and LSTM baselines. Of special note, the "Just Listen" (low engagement) condition continued to prove difficult for all models, indicating the necessity of improved feature design or multimodal fusion.

EEG-based models showed robust performance with Catch-22 attributes, although deep learning approaches were not tested because of computational issues. Notably, pupillometry proved to be a promising and less invasive option compared to EEG, with comparable accuracy but better applicability to wearable and real-time systems. The results point to the compromise between accuracy and latency, highlighting the need for light, interpretable models for real applications.

Subsequent research will focus on applying pupillometry and EEG analysis to mental health applications, especially depression diagnosis, where atypical cognitive load responses can yield informative biomarkers. Hybrid and attention-based deep models might be researched to balance efficiency with accuracy and maintain interpretability. Fusion strategies in multimodality should also include pupillometry as the main signal and EEG as a supplementary channel in high-end diagnostics. Lastly, deploying the suggested models on wearable devices like spectacle-mounted eye trackers will allow for constant, real-world observation, thus promoting large-scale longitudinal neuroscience and mental health studies.

\section{Code Availability}
The code and experimental resources used in this study are publicly available at:
\url{https://github.com/NusaibahFarrukh/PupillometryBrainDynamics}

\section{Acknowledgements}
The authors would like to extend their sincere thanks to the supervisor, project mentor and the faculty for their guidance. We acknowledge the OpenNeuro platform for providing the dataset and the computational resources provided by the University’s GPU server facility.. Large Language Models (LLMs), were used only for language editing, proofreading, and formatting assistance. No part of the methodology, data analysis, results, or scientific content was generated by LLMs and all core research ideas, experiments, and interpretations are the authors’ own.

\end{document}